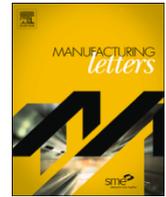

52nd SME North American Manufacturing Research Conference (NAMRC 52, 2024)

# Towards a Digital Twin Framework in Additive Manufacturing: Machine Learning and Bayesian Optimization for Time Series Process Optimization

Vispi Karkaria[a], Anthony Goeckner[a], Rujing Zha[a], Jie Chen[a], Jianjing Zhang[b],

Qi Zhu[a], Jian Cao[a], Robert X. Gao[b], Wei Chen[a]

[a]*Northwestern University, Evanston, IL 60208, USA*
[b] *Case Western Reserve University, Cleveland, OH 44106, USA*

\* Corresponding author. Tel.: +1-224-766-1920;. E-mail address: weichen@northwestern.edu

**Abstract**

Laser directed-energy deposition (DED) offers notable advantages in additive manufacturing (AM) for producing intricate geometries and facilitating material functional grading. However, inherent challenges such as material property inconsistencies and part variability persist, predominantly due to its layer-wise fabrication approach. Critical to these challenges is heat accumulation during DED, influencing the resultant material microstructure and properties. Although closed-loop control methods for managing heat accumulation and temperature regulation are prevalent in DED literature, few approaches integrate real-time monitoring, physics-based modeling, and control simultaneously in a cohesive framework. To address this, we present a digital twin (DT) framework for real-time model predictive control of process parameters of the DED for achieving a specific process design objective. To enable its implementation, we detail the development of a surrogate model utilizing Long Short-Term Memory (LSTM)-based machine learning which uses Bayesian Inference to predict temperatures across various spatial locations of the DED-built part.  This model offers real-time predictions of future temperature states. In addition, we introduce a Bayesian Optimization (BO) method for Time Series Process Optimization (BOTSPO). Its foundational principles align with traditional BO, and its novelty lies in our unique time series process profile generator with a reduced dimensional representation. BOTSPO is used for dynamic process optimization in which we deploy BOTSPO to determine the optimal laser power profile, aiming to achieve desired mechanical properties in a DED build. The identified profile establishes a process trajectory that online process optimizations aim to match or exceed in performance. This paper elucidates components of the digital twin framework, advocating its prospective consolidation into a comprehensive digital twin system for AM.

## 1. Introduction

Laser directed-energy deposition (DED) is an additive manufacturing (AM) technique where powder or wire feedstock is introduced into a laser-formed moving melt pool, creating a deposition bead. Stacking these beads produces a three-dimensional part. Laser DED allows for near-net shape fabrication of freeform geometries with internal features and offers material grading unattainable by conventional methods, facilitating localized material optimization [1]. However, broad application of DED is currently limited by heterogeneity and anisotropy in DED part mechanical properties, as well as part-to-part variation, due to the incremental nature of DED fabrication [2].

DED's layer-wise fabrication, paired with high melt temperatures, induces cyclic thermal cycles, leading to unique microstructural textures  [3]. The cumulative heat and additive nature create uneven thermal histories. Although post-build treatments can moderate this, some DED phenomena, like δ-phase growth in Inconel 718, optimize subsequent treatments [4]. Thermal accumulation can result in chemical alterations due to selective evaporation of elements, while heterogeneous cooling conditions can result in spatially variate mechanical properties [5,6]. While hybrid additive manufacturing that introduces secondary energy sources can adjust microstructure, they increase costs and complexity [7]. Control of DED properties via control of laser power therefore remains a worthwhile technique. Thermal objectives may be derived from experimentally observed insights, or solidification simulations using phase-field, cellular automata, or kinetic Monte Carlo techniques [8].

Several works in DED sought to minimize part variations, using direct melt pool temperature measurements as an input for a PID controller. For instance, a rule-based controller was used to stabilize melt pool temperatures, reducing part-to-part porosity variation [9]. Such control schemes fail to take advantage of disturbances known *a priori*, while feedforward control can proactively anticipate and compensate for such disturbances. For instance, Liao *et al.* employed a virtual PID in a heat transfer simulation to generate a feedforward control input [10]. Integrating these with feedback control compensates for discrepancies in the model versus real-world conditions. A hybrid feedback-feedforward system, merging a heat transfer model with a PI controller, effectively maintained melt pool temperatures by adjusting laser power [11]. However, such models may not intelligently track changes in system dynamics if such dynamics were not modelled explicitly. Incorporating additional physics to these closed-loop controllers typically increases the cycle time, reducing controller performance.

Evidently, there is need for a manufacturing system model for DED that 1) models a target parameter, such as temperature or flaw concentration in the relevant regions that cannot be directly observed, 2) is updated during the online process to account for differences between the physical system and the virtual model, and 3) communicates changes in the system state to the real system, aiding control decisions based on these updated conditions. Such a system is termed a digital twin (DT) [12,13]. Though many works have built so-called "digital twins" for DED that track model dynamics as a response to in-situ measurement data, these predictions are often not suited for online control due to the computational complexity and the lack of the characteristic bidirectional communication between the digital and physical twins. As a result, these are more suited to be termed as "digital shadows" only.

To build this digital twin for part temperature in DED, we must first construct a fast surrogate model for online evaluation. Various methods have been used for rapid, reduced-order temperature models in AM. Physics-informed neural networks [14] were used to assess transient temperatures near a DED-like heat source [15]. However, modifications in boundary conditions, such as geometry changes, require extremely expensive retraining. GRU neural networks [16] and later, recurrent graph networks [17], were used for temperature prediction in DED parts. While they perform well for new geometries and longer time steps, their effectiveness under dynamic laser power variation remains untested. The heat conduction graph network [18] accommodates variable laser power, but its predictions lack the uncertainty quantification (UQ) and real-time prediction which is vital for the digital twin framework. UQ offers insights into model accuracy, a required feature for intelligent decision-making in the online system.

Moving forward the offline trained surrogate model should be employed for online predictive control. In most cases, online predictive models need a process trajectory, which is obtained by offline optimization. As accurate DED simulations can involve detailed physics, they are typically expensive to execute. It is therefore beneficial to minimize the number of iterations in the optimization process [19]. Attempts at finding an optimized laser power profile of DED have been made; however, many require a fully differentiable surrogate model [20]. Differentiable simulations make the derivatives of objective functions with respect to process inputs directly available; however, they have very high memory requirements, which limits their applicability to part scale builds.

Recognizing these challenges, the integration of digital twin technology within the laser DED process aligns with Industry 4.0's drive towards intelligent manufacturing [21]. Digital twins, key elements of Industry 4.0, facilitate a dynamic manufacturing environment, enhancing real-time control and predictive accuracy through seamless integration of cyber-physical systems [22]. This approach not only embodies the principles of Industry 4.0 but also marks a significant advancement in adaptive, data-driven manufacturing process control and optimization.

In this study, we present a digital twin architecture to control and optimize manufacturing processes, with the following contributions:

1. We propose a digital twin framework for the additive manufacturing process that facilitates a bidirectional information exchange between virtual and physical systems where the framework incorporates online model predictive control for real-time process adjustments and offline model updates for enhanced prediction accuracy. Additionally, the framework mitigates sources of uncertainties including the "unknown of unknowns".
2. We develop a Bayesian LSTM architecture as a surrogate model which predicts the future temperature state by looking at the past measurable temperature state, and other input variables It handles the spatial-temporal dynamics of the DED process and emphasizes significant past events while quantifying uncertainty—vital for informed decision-making within a digital twin framework. Different from conventional LSTMs, our model adopts Bayesian inference, utilizing Monte Carlo dropout methods to gauge predictive uncertainty by estimating Bayesian posterior distributions.
3. We develop a low-dimensional representation for high dimensional time series of process inputs by decomposing high-dimensional laser profile time series into a set of low-dimensional parametric time series functions that capture distinct laser profile characteristics. This enables the use of BO to find an optimal time series profile with adaptive sampling using fewer samples. We demonstrate the capabilities of this method by optimizing a time series profile of laser power with the objective of maximizing heat treatment time throughout DED-built parts for better material quality.

This paper is structured as follows: Section 2 delineates the architecture of a digital twin framework applied to additive manufacturing. Section 3 discusses the implementation and validation of a Bayesian LSTM designed to forecast

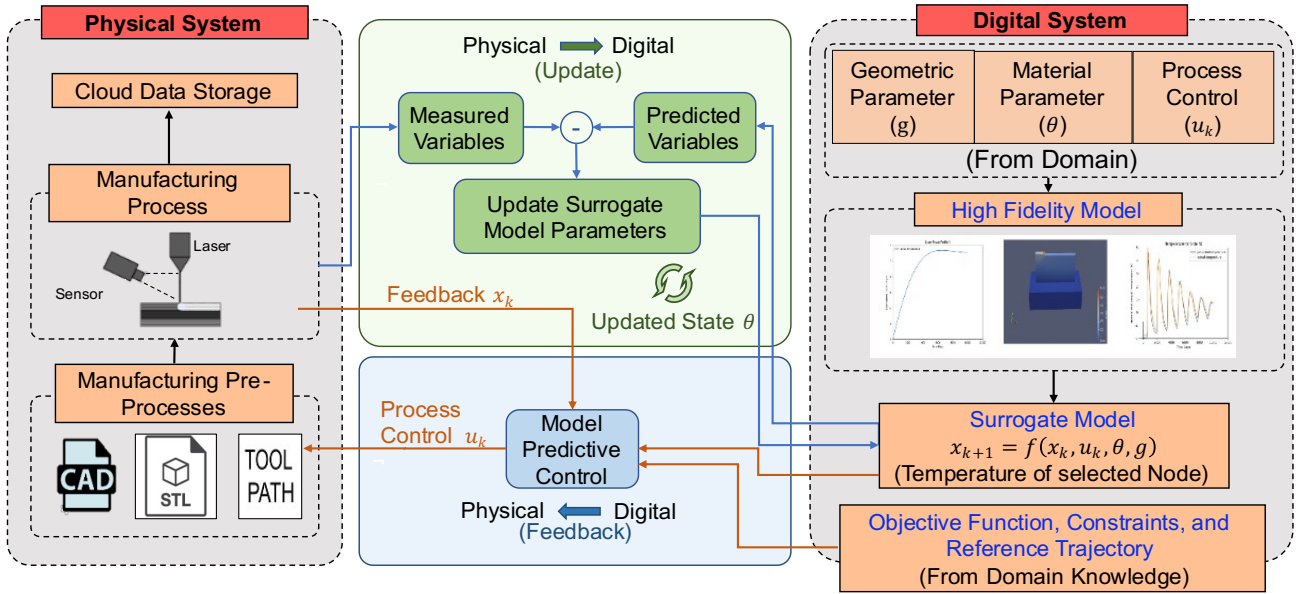

Fig. 1. Schematic of our proposed Digital Twin Framework for DED Additive Manufacturing; In this paper we focus on the Bayesian LSTM, the Surrogate Model we selected, and the offline optimization, which aids the Model Predictive Control

temperatures during laser DED. Section 4 explicates the novel Bayesian Optimization for Time Series Process Optimization (BOTSPO) method, which reduces the complexity of time-series data for laser power inputs in DED processes and examines its efficacy in deriving an optimal power profile. The paper concludes with a synopsis of its contributions and a discourse on prospective research directions.

## 2. Digital Twin (DT) framework for the Additive Manufacturing Process

Many existing DTs only provide a "digital shadow" of the physical system. In this work, a DT framework is proposed to integrate physics-based models together with real-time data collection (physics-based ML) for fast and accurate -property prediction and process optimization. Different from existing works that use system performance as objective function in optimization or decision making, we introduce two-way (physical-to-virtual and virtual-to-physical) information-centric value proposition. Model Predictive Control (MPC) is included for anticipating and taking control actions based on current and future events, which distinguish it from linear–quadratic regulators and PID controllers. Unlike standard practices where model validation is a one-time, offline event, our DT leverages continuous online data to ensure the model remains accurate and reliable, effectively bridging the gap between the physical and digital realms. This method advances how we implement model predictive control, offering a more adaptive and responsive system compared to conventional control strategies.

To demonstrate the above-described DT framework, we present a DT to support real-time DED manufacturing process optimization, which is shown in Fig. 1. The left block is the physical system, and the right block is the digital system. The two blocks in between show the bidirectional interplay between the physical system and the digital system. The physical system is the manufacturing process which is guided and programed by pre-process inputs and monitored by sensors. The process parameters are controlled, e.g., laser power. To achieve the objective of temperature control, the digital system is created using high fidelity finite element method (FEM) simulations which take geometric parameters, material parameters and process parameters as inputs, and predict temperature at each location of the built part. In this study, the process control parameters are the laser power at each step in time. To enable real time analysis and decision making, a surrogate time series model is needed to replace the expensive FEM simulations. Uncertainties exist in the DT including aleatoric uncertainty arising from inherent randomness and unavoidable variations (e.g., material and manufacturing variability, and sensor noise) and epistemic uncertainty arising from limited knowledge and data (e.g., numerical uncertainty, and model uncertainty). To account for both aleatoric and epistemic uncertainties, the predictive model is updated by combining data from both the physical and the digital system. The offline updated surrogate model is then fed into an online MPC system. Different from traditional feedback control techniques, MPC can better handle constraints and nonlinearities associated with the object or system of interest and performs process optimization to identify the best control action based on the current state of the system, a predictive model, and a decision-making objective. MPC optimizes the current timeslot while considering future timeslots through iterative optimization of a finite time-horizon, implemented sequentially. With real-time sensing, the capability of MPC allows the digital twin to make real-time (online) adjustments and optimizations in response to uncertainty and variations that cannot be predicted or mitigated offline, the so-called unknown of unknowns (i.e., unexpected or unforeseeable conditions).

In this work, we focus on the development of the surrogate model and offline process optimization in the digital system, which are described in the following sections.

## 3. The Bayesian LSTM network as a surrogate model

### 3.1 Architecture of the Bayesian LSTM model

To be effective for real-time operation, and tightly coupled operation with a physical system that it mirrors, the digital twin must be capable of rapid data processing and control [12]. As even simplified pure physics models are much too slow to run in real time, we turn to a machine learning (ML) model to estimate the temperature throughout the part [23]. We achieve this speedup using a recurrent neural network-based surrogate model, which also considers past events while making predictions. In this section, we describe the architecture of the model and the design decisions behind it.

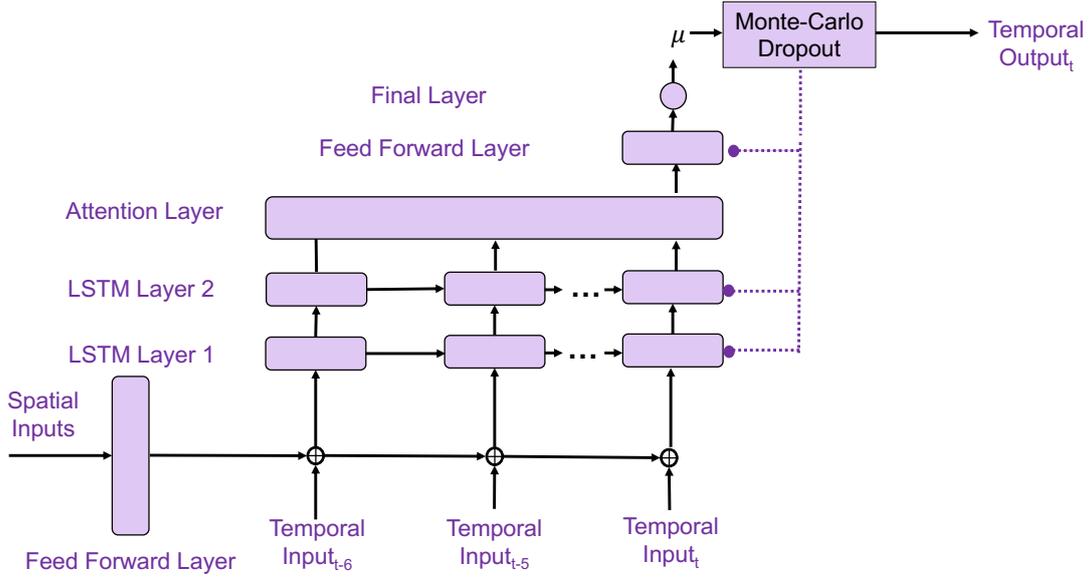

Fig. 2. Architecture of the LSTM-based neural network to predict temperature at different spatial locations; time-invariant inputs include laser power at element birth ($LP_{birth}$), and Node deposition time ($T_{birth}$), while time-varying inputs include distance from node to edge of built part (closest) ($DN_t$), distance from laser deposition point to node ($DL_t$), laser power at current time ($LP_t$), and temperature of previous time step at the same location ($T_{t-n}$). The temporal output is the temperature of the current time step of the selected node.

The surrogate model as shown in Fig. 2 predicts the temperature history at a particular point by taking the following features, as illustrated by Fig. 4: the distance from the point to the laser deposition point ($DL_t$) and to the nearest surface boundary ($DN_t$), the time at which the point is deposited ($T_{birth}$), the laser power at the deposition time ($LP_{birth}$), the current laser power ($LP_t$), and the measured temperature from the previous time window ($T_{t-1:t-n}$). We identify these inputs from previous successful RNN-based modeling of the DED process [16], with the addition of the time and spatially variant laser power features. The current temperature prediction at a location can be written as a function of Bayesian LSTM model as follows:

$$T_t = f(DL_t, DN_t, T_{birth}, LP_{birth}, LP_t, T_{t-n}) \qquad (1)$$

These input features are either time-variant or time-invariant. We found through experimentation that the Bayesian LSTM performed poorly when provided with time-invariant data as opposed to the time-variant data. This performance degradation occurs even if only a portion of the input data is time-invariant. Therefore, our surrogate model is composed of two primary data paths, one for time-variant data and the other for time-invariant data, which are eventually concatenated. The time-invariant data is fed to a forward layer with 100 nodes whose output corresponds to spatial information, such as location.

In this model, we have two LSTM layers with 500 LSTM neuron units. This neuron count strikes a balance between computational efficiency and model complexity. LSTMs are preferred in our study for their architecture, which includes a memory cell and three gates. This enables the model to effectively regulate the flow of information, remembering patterns over long sequences and discarding irrelevant data [24]. The selection of these hyperparameters followed a tuning process to ensure the model's performance is maximized for our specific dataset and task requirements. Their ability to handle complex dependencies makes them similar to GRUs for modeling the intricate thermal patterns in DED applications [25,26]. Their capability to utilize past data for future forecasts aligns with our need to monitor the evolving state of the DED-built part [27]. Our enhanced Bayesian LSTM model integrates an attention layer [28]. Specifically, the attention mechanism in our Bayesian LSTM model operates by assigning varying weights to different time steps in the temperature data sequence, thereby allowing the network to 'attend' more to moments with significant thermal activity and less to others [29,30]. This facilitates the model's ability to dynamically emphasize temperature anomalies or shifts that could lead to defects, enabling precise adjustments to process inputs.

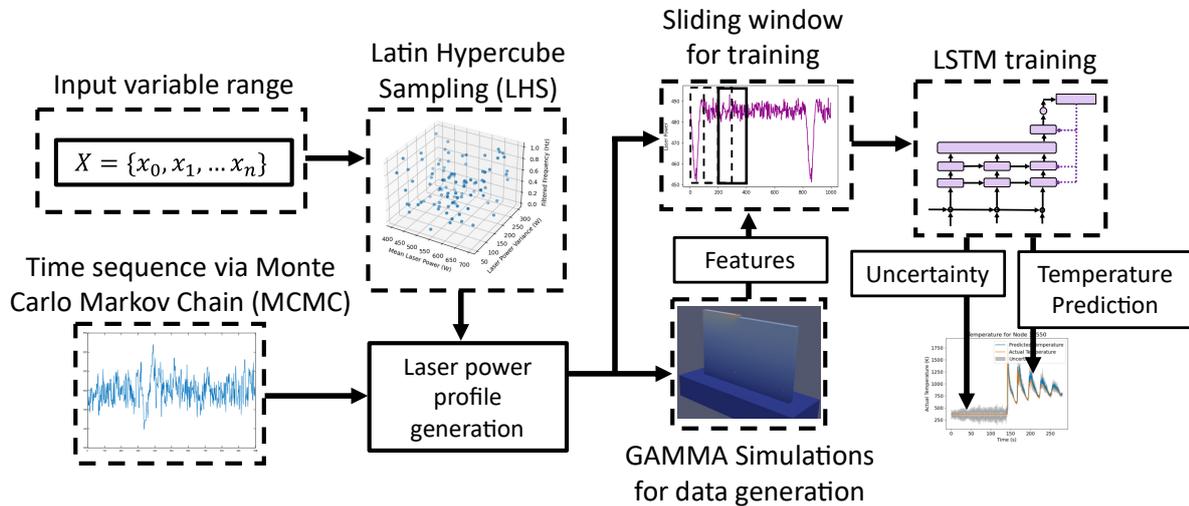

Fig. 3. Flowchart demonstrating the data preparation, simulation, and training strategy steps of section 3.2 "Bayesian LSTM model training strategy"

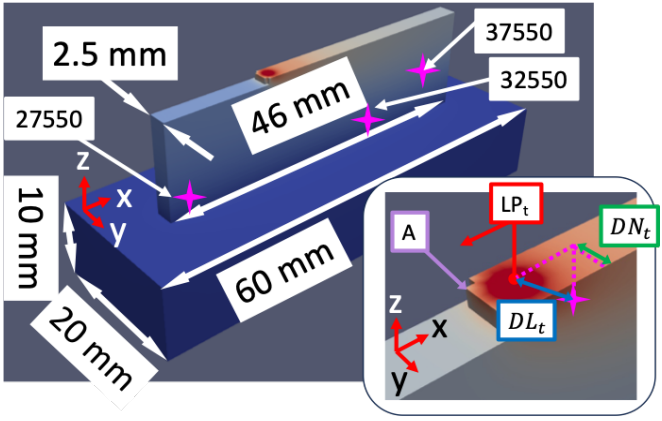

Fig. 4. Snapshot of FEA simulation, illustrating extracted features. The laser, indicated by $LP_t$, travels in the direction of the red arrow. The point of interest is represented by the pink star, which is at a distance $DN_t$ from the closest free surface and $DL_t$ from the laser. The laser power is recorded for the current time, as well as the time at which the first element containing the node was activated. When an element in the current layer is within the laser spot boundary, (A in the Fig.), it is activated. The final wall height is 30 mm. The location of the three nodes under consideration are included.

To enhance the model's accuracy in representing the spatial variations in DED, we incorporated a feedforward layer with 100 nodes [31]. This feedforward layer aggregates the features extracted by the LSTM layers, combining the sequential temporal data processed by the LSTM with spatial correlations [29] Recognizing the importance of model robustness, we introduced post-LSTM dropout layers. Dropout techniques [32], serve as effective regularization tools that prevent overfitting. Our model employs Monte Carlo dropout to quantify uncertainty alongside mitigating overfitting by sampling with random node exclusion. Specifically, Monte Carlo dropout is applied during both the training and prediction phases to all the layers of the neural network, repeatedly 'turning off' a random subset of neurons in the network [33]. This process creates multiple 'thinned' networks, each generating its own predictions [34]. The variance among these samples indicates uncertainty. Through this architecture, our model can predict future temperature while simultaneously quantifying uncertainty.

*3.2 Bayesian LSTM model training strategy*

To train our Bayesian LSTM machine learning model, we used FEM simulations to generate a bank of simulated full-field temperature measurements for our test geometry. To that end, we used a GPU-accelerated explicit heat transfer model to simulate our test geometry's temperature history across a range of deposition laser power profiles. Then, we sampled temperature histories from individual locations across the build, and used the temperatures experienced by these individual points to train the data-driven model. This approach is illustrated by Fig. 3.

Fig. 4. illustrates the test geometry used in this work, which consists of a bidirectionally scanned thin wall constructed from Inconel 718 built on a substrate of low-carbon 1018 steel. The scanning speed was set to 7 mm/s, and the $1/e^2$ beam diameter was set to 2.24 mm. The build consists of 40 consecutive 0.75 mm tall layers. The simulation parameters are calibrated to an in-house laser-powder DED machine, the Additive Rapid Prototyping Instrument (ARPI). A detailed discussion of the formulation of the thermal model and comparison of full-field temperature predictions with the physical DED system be found in Liao *et al*. [10]. The model was also proven to offer sufficiently accurate melt pool temperature predictions for laser power optimization in [11].

A two-step method, combining Latin Hypercube sampling (LHS) and Markov-chain Monte Carlo (MCMC) was used to generate a representative library of laser power profiles; namely, LHS is used to sample from a space of parameters that describe the laser history. LHS is used for its ability to efficiently sample from the entire high-dimensional space of experimental parameters [35]. To accurately sample the space of possible laser profiles, the laser power profile was broken into three fundamental properties: frequency, mean, and variance. The frequency component captures how rapidly the laser power is allowed to change from one time step to the next. High-frequency changes are reflective of events such as rapid drops in laser power, as may be experienced in an optimized trajectory near a turnaround point [11]. The mean component alters the average of the laser power in the trajectory. Finally, the variance component gauges the extent of variability or fluctuation within the laser power trajectory. These three variables provide a comprehensive coverage of the parameter space [36].

Using LHS, 50 parameter sets were generated within set bounds (Mean: 400-700 W, Variance: 50-300 W, Frequency: 0-1 Hz) heuristically based on prior IN718 deposition experience. The profiles were then generated via MCMC in arbitrary unit space and scaled and shifted to the requested mean and variance. The signals were then filtered to the required temporal frequency with a 2nd-order low-pass Butterworth filter. The Butterworth filter is a general-use filter in signal processing with a flat frequency response in the passband [37].

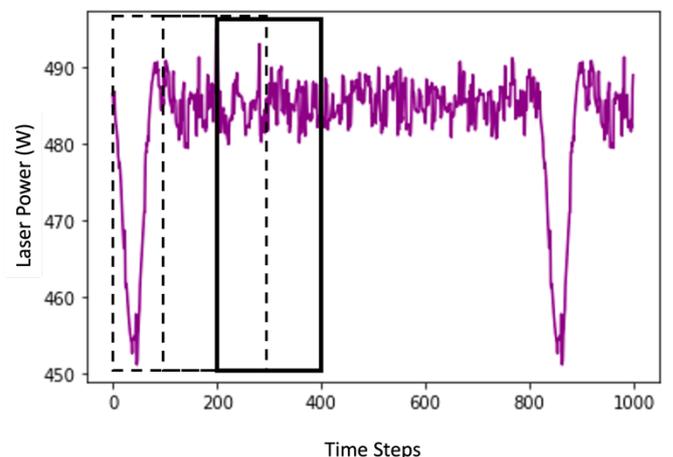

Fig. 5. Schematic of sliding window which is applied to feed data while training the Bayesian LSTM Model. The sliding window has a length of 200-time steps, which is 2 seconds. The frequency is 50 Hz, and the sliding window slides for 100 time steps.

From each run of the FEM simulation, we collect various data which will form the input features to our surrogate model. Throughout the paper, we refer to "locations" of importance in the DED-built part; these locations are one-to-one mappings with nodes in the finite element model. At each time step, we record the input and output data required to train our machine learning model. The finite element simulation runs at an explicit time step of 0.002 seconds. The full-field temperature field and training parameters are continuously gathered every 10 simulation steps, or every 0.02 seconds. We then repeat the simulation using different laser profiles, resulting in training datasets for each laser profile.

To achieve location-agnosticism of the surrogate model, we must sample from datasets of multiple laser profiles from all the available 96000 nodes during training. For this purpose, we have developed a data management pipeline. Each sample used for training is a sliding window of time-series data [38] as shown in Fig. 5. During sampling, time windows are overlapped such that the latter half of data points in one window are also included as the former half of data points in the next window [39]. During training, individual time windows are sampled randomly from datasets of 80 laser power profiles of 280 seconds to create mini batches of samples.

Using these mini-batches, training of the model proceeds using the gradient descent method. Loss is calculated using the evidence lower bound (ELBO) [40] method and is defined as follows:

$$L(\phi, \theta, x) \coloneqq \mathbb{E}_{z \sim q_\phi(\cdot|x)}[\ln p_\theta(x|z)] - D_{KL}(q_\phi(\cdot|x) \parallel p) \quad (2)$$

where $x$ is the observed data points, $\phi$ are the parameters defining the approximate posterior distribution, $\theta$ is the Parameters of the true intractable distribution, $z$ is a latent (unknown) continuous random variable, $p_\theta(x|z)$ is the true, intractable distribution of $x$, $q_\phi(\cdot|x)$ is the approximate posterior, and $D_{KL}$ is the Kullbach-Leibler (KL) divergence from distribution $p_\theta(\cdot|x)$ to distribution $q_\phi(\cdot|x)$. The ELBO loss function is originally described in Auto-Encoding Variational Bayes [40], and we found via experimentation that it is highly effective when used for training our surrogate model. The ELBO method stands out for uncertainty quantification due to its decomposition of the marginal likelihood into two interpretable components: the expected log-likelihood and the KL divergence. The first term, expected log-likelihood, quantifies how well the model explains the observed data, whereas the KL divergence penalizes deviations from the prior and acts as a regularize, preventing overfitting. This decomposition provides a robust balance between data fidelity and model complexity, ensuring a principled quantification of uncertainties. In the next section, we will look at the prediction capability of the Bayesian LSTM model.

*3.3 Evaluation of the LSTM Model*

In evaluating the performance of our Bayesian LSTM model, we compare the resulting outputs against simulated measurements. In our first evaluation we input the past measurable temperature or the actual temperature as an input. The LSTM model's performance assessment is carried out using a set of 20 distinct laser power profiles, each with a duration of 280 seconds, specifically reserved for validation purposes.

In this case the models achieved an $R^2$ score of 0.75, indicative of a good degree of accuracy and alignment with the expected outcomes. The $R^2$ score, commonly referred to as the coefficient of determination, quantitatively measures the proportion of the variance in the dependent variable that is predictably explained by the independent variables in a regression model. This result was validated through k-fold cross-validation and independent set evaluation to ensure model robustness against overfitting.

The temperature profiles for nodes 27550, 37550, and 32550 are shown over an extended period for a representative laser power profile in Fig. 6. These points were chosen to illustrate the thermal history at a variety of distances from the turnaround regions and part surface. Fig. 6. shows that the ground truth temperature (represented in orange) aligns closely with the LSTM predictions (shown in blue). The accompanying shaded region illustrates the uncertainty bounds, which are derived using a 95% confidence interval. By examining these diagrams, it is evident that while the model offers a reasonable approximation of the actual temperature profiles, the prediction quality is poorer when the laser is further from the node under investigation, with predictions tending to be higher than the ground truth simulation.

For effective prediction of the internal temperatures of the fabricated part—temperatures which are inherently unobservable—the model can leverage prior temperature estimations from the Bayesian LSTM to subsequently predict upcoming temperatures. This sequential prediction will be important when interfaced with the physical plant. While the main emphasis of this study revolves around surface temperatures, where prior temperatures are assumed to be known, the LSTM may also be run using the temperature predictions at previous time steps as input. As shown in Fig. 7, the model reliably predicts internal temperatures with a confidence level of ninety-five percent for six seconds. Although the model demonstrates high predictive accuracy overall, the model's reliance on its own outputs results in substantial error accumulation past the first temperature peak.

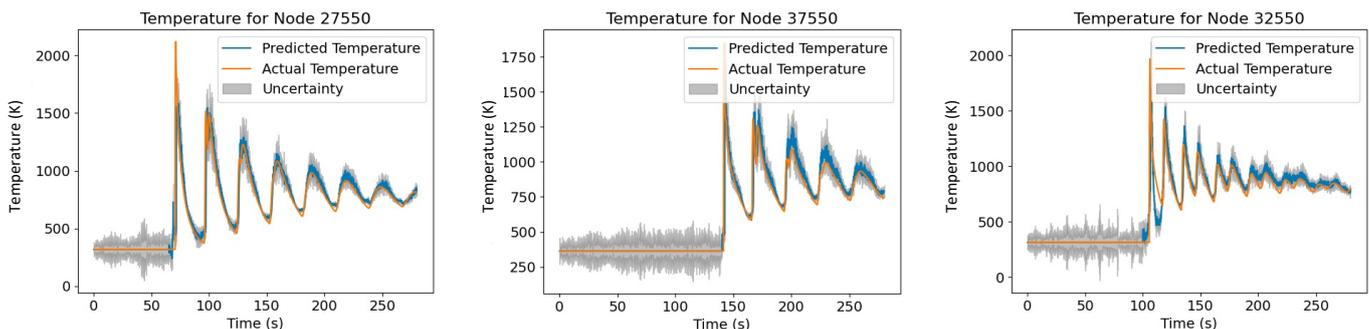

Fig. 6. Evaluation of the Bayesian LSTM model on test dataset for three locations in the thin wall with a sliding window prediction length of 2 seconds in future over a total duration of 280 seconds; Comparison of model prediction (blue line) and the actual value (orange line) with 95% uncertainty band (grey shaded area).

In the digital twin context, the Bayesian LSTM model serves as a critical predictive tool for advanced manufacturing. By incorporating the Bayesian LSTM into the digital twin framework, the system can proactively monitor thermal behavior in real-time during DED processes and update its understanding of system dynamics. This integration is vital for precision in complex structures, where slight changes in deposition conditions can greatly affect the properties of built parts.

## 4. Bayesian Optimization for Time Series Process Optimization (BOTSPO)

*4.1 Description of the BOTSPO method:*

Digital twin technology underlines the essential link between offline analyses and online implementations. Offline optimization fine-tunes parameters using historical data to enhance online predictive control effectiveness [41,42]. Fundamentally, offline optimization transcends mere preparatory operations, laying the foundation for the performance of online optimization [43].

In this section, we develop a methodology to efficiently represent the laser power profile utilizing a minimal set of parameters as shown in Fig. 8. Within the BO framework, we employ a Gaussian process (GP) model to capture the nonlinear relationships between laser profile parameters and the desired characteristics of laser profile along with their uncertainties, thereby enabling robust and efficient exploration and exploitation of the process design domain. In BO, as the dimensionality of the parameter space increases, the computational complexity of the GP inference grows exponentially, leading to prohibitive computational overheads [44]. This phenomenon, often referred to as the "curse of dimensionality," necessitates judicious selection of parameters to ensure the optimization remains tractable and efficient. We developed a Fourier series-based time series profile generator for laser power, leveraging its ability to dissect signals into sinusoids for clear frequency domain analysis and reconstruction of complex laser behaviors.

To derive our time series generator, we start with a simple Fourier series profile [45]. Fourier series offer immense flexibility because they can approximate any periodic function with a weighted sum of sines.

$$y(t) = A \left( \frac{2}{\pi} \sum_{i=1, i \bmod 2=1}^{n} \frac{1}{i} \sin(2\pi \cdot f \cdot i \cdot t + (\phi)) \right) \quad (3)$$

In the above Fourier series formula, the parameters $A$, $n$, $f$, and $\phi$ represent the amplitude, number of terms, frequency, and phase of the Fourier series respectively. To keep the number of parameters low, and to increase the flexibility of the time series profile generator we introduce $\Delta A$, $\Delta f$ and $\Delta \phi$.

$$y(t) = (A + n\Delta A) \left( \frac{2}{\pi} \sum_{i=1, i \bmod 2=1}^{n} \frac{1}{i} \sin(2\pi(f + i\Delta f)it + (\phi + i\Delta\phi)) \right) \quad (4)$$

In the above time series generator, we employ a modified Fourier series characterized by three parameters: $\Delta A$, $\Delta f$, and $\Delta \phi$. The $\Delta A$ parameter represents the rate of change of amplitude, adding adaptability to the model by allowing the amplitude to evolve over time. Concurrently, $\Delta f$ signifies the rate of change of frequency, enabling the model to responsively adjust the oscillation frequency of the laser power. Lastly, $\Delta \phi$ describes the rate of change of the wave's phase, ensuring the model can compensate for temporal shifts in the laser power profile for alignment with real-world observations. The introduction of these rates of change (Δparameters) provides the representation with the flexibility to change its morphology gradually during the build while only introducing a three new terms.

Temporal trends and seasonality are essential components when modeling time series data, which comprises data points listed at successive time intervals. A temporal trend, signified by the term $t$ in the equation $y'(t) = y(t) + T \cdot t$, denotes a consistent trajectory the data adheres to over time. The coefficient $T$ represents the slope, capturing whether the laser power profile gradually increases, decreases, or remains stable over time. While the Fourier terms could theoretically capture such a trend, explicitly implementing a linear trend reduces the burden on the sinusoidal terms to represent noncyclic behavior.

On the other hand, seasonality captures recurring local behavior in a time profile. This cyclic behavior is embodied in the term $S\sin(2\pi \cdot t \cdot SF)$ in the equation. Here, $S$ adjusts the amplitude of the seasonal fluctuation, SF represent the frequency in which this cyclicity is captured, and the sine function represents the cyclical pattern. Seasonality can also be captured by a custom function if the underlying cyclic behavior is known. By integrating both the temporal trend and seasonality, the equation $y'(t)$ provides a comprehensive portrayal of the laser power profile over time, ensuring the digital twin closely mirrors real-world dynamics.

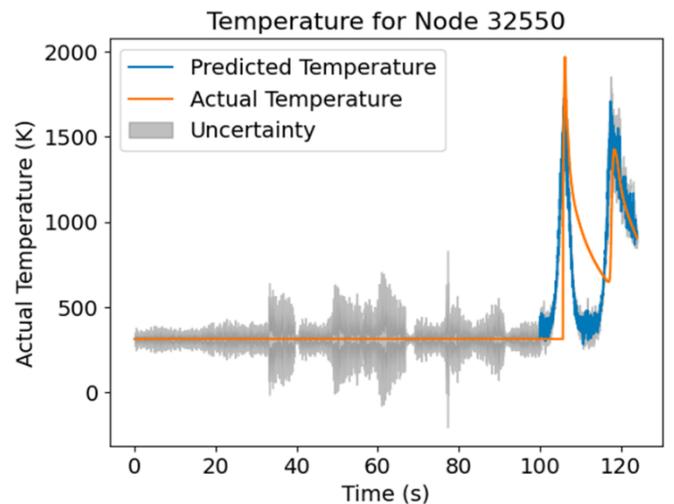

Fig. 7. Evaluation of Bayesian LSTM with its own temperature prediction as an input to the LSTM

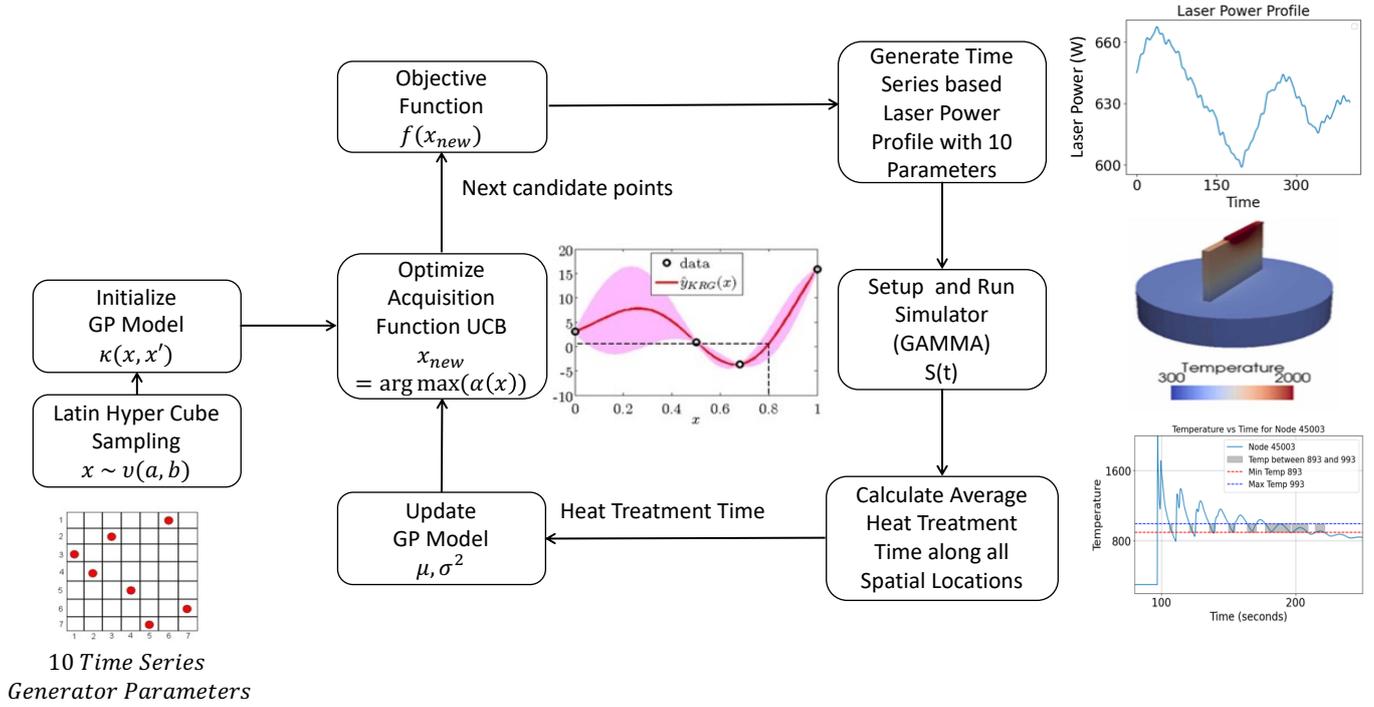

Fig. 8. Framework illustrating the data flow in Bayesian Optimization for Time Series Process optimization (BOTSPO). This application of BOTSPO is specifically illustrated for finding the optimal laser power for DED.

$$y'(t) = y(t) + Tt + S\sin(2\pi t \cdot SF) \qquad (5)$$

In total, the system utilizes ten tunable parameters, encompassing amplitude, frequency, phase, their rates of change, and temporal dynamics to accurately capture and represent the laser power profile over time, yielding a robust and adaptive model for real-world applications.

The selection of ten parameters for the time series generator was empirically determined to strike a balance between model complexity and computational efficiency. These terms allow for a sufficiently detailed representation of the laser power profile (time series profile), capturing the essential dynamics without overfitting, which is a common concern with more parameters [46].

To optimize the time series laser power profile, we employed BO, a model-based optimization approach for maximizing GP Models. The GP Model function links the ten adjustable parameters of our time series to the target objective. Denoting the ten parameters as $p$ where $p = [p_1, p_2, ..., p_{10}]$ and the heat treatment time as $f(p)$, the GP Model will be,

$$f(p) \sim GP(m(p), k(p, p')) \qquad (6)$$

where $m(p)$ is the mean function and $k(p, p')$ is the covariance function or kernel. In our study, we employed the radial basis function (RBF) [47]. The Radial Basis Function (RBF) kernel is a covariance function used in Gaussian processes that measures similarity based on the Euclidean distance between points in the input space, with points closer to each other having higher covariance. It is recommended to set an upper limit, and lower limit of the ten parameters of the time-series profile generator while linking the time-series profile generator to the BOTSPO. The time-series profile generator can generate a wide range of time series profiles which might not be required for all the applications, and this limit on the parameters should be made according to domain knowledge.

Our optimization sequence, as outlined in Fig. 8, initiates with Latin hypercube sampling to establish an initial Gaussian Process (GP) model, setting the stage for capturing the underlying function dynamics of the DED. Under this setup, the Latin hypercube sampling method is employed to efficiently cover the 10 dimensional parameter space. This initial GP model then serves as a probabilistic surrogate for the objective function, facilitating the estimation of both the function's value and uncertainty across the parameter space. Then, BO focuses on the objective function and leverages the ten parameters to generate a respective time series laser profile. Following the initialization phase, the BO process exploits the GP posterior to identify regions in the parameter space that likely contain optimal solutions. To make this selection, we employ the Upper Confidence Bound (UCB) as our acquisition function, formulated as:

$$UCB(x) = \mu(x) + \kappa\sigma(x) \qquad (7)$$

where $\mu(x)$ is the GP posterior mean, $\sigma(x)$ represents the standard deviation, and $\kappa$ is an exploration-exploitation parameter that dictates the trade-off between exploring uncertain regions and exploiting known optima. By adjusting $\kappa$, we can finetune the balance between exploration and exploitation in our optimization routine. As iterations progress, the GP model is updated with the newly observed data, refining the predictive landscape. The BO iteratively utilizes the UCB acquisition function to select new samples, specifically targeting regions that promise the most significant potential improvement. It guides the search towards an optimal laser power profile by aiming to maximize the UCB with respect to the objective function as shown in equation (10) that is characterized by the ten parameters.

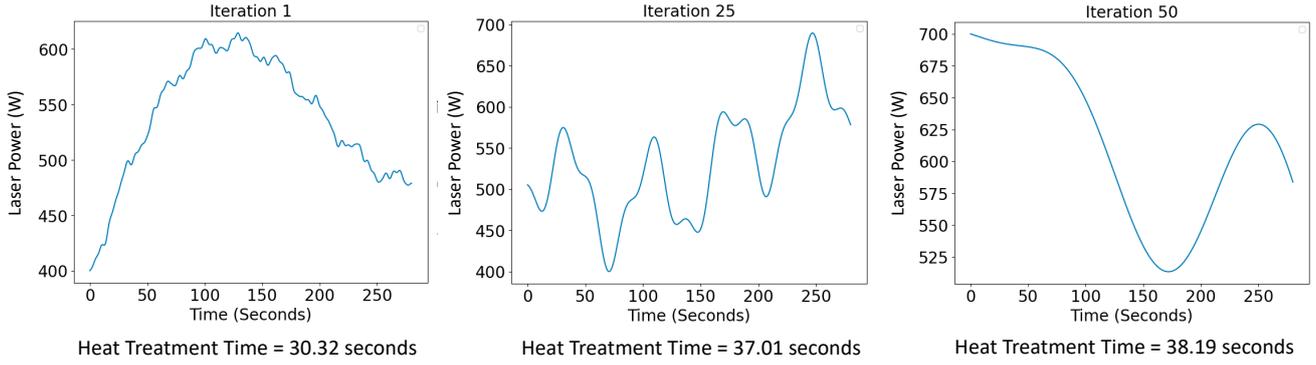

Fig. 9. Laser Power Profiles at different BOTSPO iterations whose aim is to maximize heat treatment time; First figure is the laser power profile at iteration 1, second figure is the laser power profile at iteration 25, and the third figure is the laser power profile at iteration 50.

To dynamically strike an optimal balance between exploration and exploitation in our BO process, we implemented an adaptive approach for the parameter $\kappa$, leveraging recent uncertainties in the GP model predictions. Specifically, we designed an adaptive $\kappa$, defined by:

$$\kappa_{adjusted}(i) = \beta_{base}(i) \cdot (1 + \alpha \cdot \rho) \quad (8)$$
$$\beta_{base}(i) = \beta_0 \cdot \gamma^i \quad (9)$$

Here, $i$ denotes the iteration count, $\beta_0$ is a base value, $\gamma$ is the decay factor fixed at 0.90, and $\rho$ calculates the standard deviation, which quantifies the mean uncertainty over the previous five BOTSPO iterations. The uncertainty adjustment factor $\alpha$ is set to 0.1. By scaling $\kappa$ with a base value $\beta_0$ and an decay factor $\gamma^i$, we ensure that exploration is more aggressive in early iterations to discover the potential regions of interest and gradually becomes more refined. The inclusion of a responsiveness factor $\alpha$ multiplied by the recent prediction uncertainty $\rho$ allows $\kappa$ to adjust dynamically, promoting a balance between exploration and exploitation based on the evolving confidence in the model's predictions [44]. By incorporating this adaptive approach, our optimization process can respond flexibly to the changing dynamics and uncertainties inherent in modeling of the DED process, providing a more responsive and efficient search for optimal laser profiles.

Striking an equilibrium between the exploration of parameter domains and the exploitation of regions of high performance is pivotal. BOTSPO's algorithmic architecture, characterized by its adaptive mechanism, inherent noise robustness, and sampling efficiency [41], is important for manufacturing process optimization when the cost of collecting data is high. This is particularly crucial for manufacturing process optimization, where the cost of data collection is high. The use of Bayesian methods allows BOTSPO to make informed decisions about which points in the parameter space to sample next, leveraging prior knowledge to improve both exploration and exploitation strategies. In the next section, we will look at the results achieved by the BOTSPO method in the context of DED.

*4.2 Results from the BOTSPO method*

For our objective function, we aimed to maximize the heat treatment time of the material used in the simulation, which is defined as follows,

$$\bar{T} = \frac{1}{N} \sum_{i=1}^{N} r \begin{pmatrix} \max\{j | T_{min} \leq df_{ij} \leq T_{max}\} \\ - \min\{j | T_{min} \leq df_{ij} \leq T_{max}\} \end{pmatrix} \quad (10)$$

Where $\bar{T}$ is the heat treatment time, $N$ is the total number of nodes, $r$ is the data collection rate, $T_{min}$ and $T_{max}$ are the minimum and maximum temperatures of the desired range, $df_{ij}$ is the temperature at the $j-th$ index for the $i-th$ node, $\min\{j | T_{min} \leq df_{i_j} \leq T_{max}\})$ is the first index where the temperature is within the desired range of the $i-th$ node, and $\max\{j | T_{min} \leq df_{i_j} \leq T_{max}\})$ is the last index where the temperature is within the desired range of the $i-th$ node.

In DED of IN718, multiple precipitates drive improved strength in IN718, including the body centered tetragonal Ni3Nb $\gamma''$ phase, which directly improves material strength [48], and the orthorhombic Ni$_3$Nb $\delta$ phase, whose plate-like morphology has been demonstrated to cut up brittle Nb-rich Laves phases, thus improving the effectiveness of post-build heat treatments [4]. Xie et al. [49], identified using data-driven techniques that the local strength of IN718 parts made using laser-powder DED was related to the time spent between 654°C and 857 °C. As typical aging times for IN718 are in the range of hours at the effective soak temperatures [50] the objective was set to strictly maximize the *in-situ* age time.

In Fig. 10, we present the BO progression, illustrating the advancement of the laser power profiles from the preliminary first iteration to the concluding 50th iteration. Initially, the GP model is trained utilizing a dataset of 50 laser power profile samples prior to its integration into the BOTSPO algorithm. This pre-training of the GP Model is imperative as it establishes a foundational understanding of the underlying parameter space, facilitating more informed and efficient exploratory decisions in subsequent optimization stages.

In Fig. 10, the BO function call history is delineated. In the preliminary iterations, there is a predominant focus on exploration to determine the optimal power profile to maximize the heat treatment time. Later stages exhibit plateaus, indicating a stable state and exploitation. A notable strength of BO, and subsequently BOTSPO, lies in its sample efficiency: it requires fewer data points relative to traditional techniques, thus achieving optimal parameter discernment at a reduced computational expense.

The parameters outlined in Table 1 control the build quality by influencing the laser power profile in the DED process. The "Amplitude of Fourier (A)" sets the base intensity of the laser; a higher amplitude translates to more energy input, essential for initial layers. The "Frequency of Fourier (f)" controls the oscillation frequency of laser intensity, affecting energy distribution over time. "Rate of Change in Amplitude (ΔA)" and "Rate of Change for Phase Shift (Δϕ)" adjust the laser's response to thermal dynamics during printing. "Linear Trend Coefficient (T)" indicates a general trend in laser power requirement as the build progresses. "Seasonality Amplitude (S)" and "Frequency of Seasonality (SF)" accounts for periodic variations in laser energy needs due to changes in the build geometry or material state. The "Rates of Change for Frequency (Δf)" adapts the laser's temporal response to the evolving thermal profile.

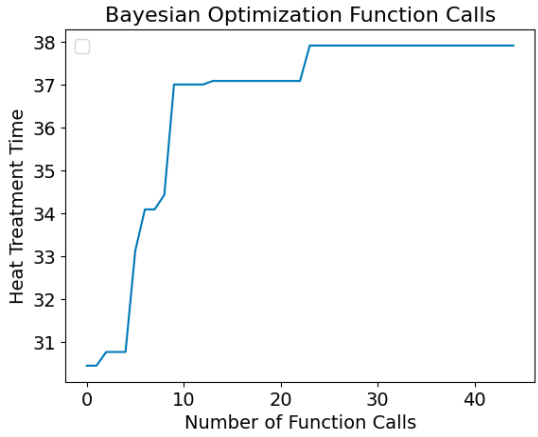

Fig. 10. BOTSPO function calls. The function call depicts the optimization history used to find the optimal laser power in this paper.

In the optimum power sequence (Fig. 9, right), the laser power is initially high. This is likely due to the longer heat treatment times at the beginning of the thin-wall build, as observed in previous works [20]. The optimizer thus sought to maximize this effect by driving up the early laser power. As the thin wall build progressed, the radiative and convective heat transfer increased; like a fin, the build became more efficient in rejecting heat and the amount of time in the heat treatment range decreased for the higher locations. The optimizer thus did not seek to increase the laser power in later parts of the build, as even laser powers at the top of the specified range did not substantially improve the total heat treatment duration.

This result, while increasing the average heat treatment duration in the part, does not achieve material homogeneity. Future work on the BOTSPO will implement multi-objective optimization that can balance joint objectives of maximizing heat treatment duration and homogeneity, thus preventing the front-biased result in this work.

Utilizing the proposed methodology allows us to determine an offline laser power profile, serving as a foundational process trajectory for online predictive control during the laser-based additive manufacturing process. This process trajectory can streamline real-time adjustments, enhancing consistent deposition quality and decreasing potential discrepancies arising from fluctuating process dynamics in laser-based additive manufacturing.

## 4. Conclusion

In this paper, we proposed a digital twin framework for DED process control and optimization, an autonomous and systematic approach that enables repeatable fabrication of complex parts. This paper contributes two major components of implementing the overall digital twin framework: a data-driven, surrogate model that enables dynamic prediction of temperatures with uncertainty quantification throughout the part in real-time, and a Bayesian Optimization Method for Time Series Process Optimization (BOTSPO) that determines appropriate laser power profiles for maximizing the part's precipitation hardening time.

First, we developed a Bayesian LSTM based surrogate model which is capable of temperature prediction for selected areas of the built part in real-time. The Bayesian LSTM model is laser profile-agnostic and is capable of accurate location-based temperature prediction for simple geometries like thin walls. We demonstrated the capability by training and analyzing the predictive capabilities of the Bayesian LSTM model on a thin wall geometry with an $R^2$ score of 0.75. This surrogate modelling capability will enable real-time MPC in the online digital system when online monitoring information becomes available. The distinctiveness of our Bayesian LSTM lies in its architecture that not only retrospectively considers measurable states and input variables but also adeptly understands the spatial-temporal dynamics inherent to the DED process. It has been designed to provide emphasis on significant historical events, while also quantifying the inherent model uncertainty—a critical feature for making informed decisions in a digital twin environment. Unlike traditional LSTMs, our model incorporates Bayesian inference principles through the implementation of Monte Carlo dropout techniques. This allows for an approximation of Bayesian posterior distributions, thereby furnishing a quantifiable measure of uncertainty in its predictions, which is important for optimizing the manufacturing process.

Table 1. Parameter Values of the Obtained Optimal Laser Power Profile at iteration 50 of BOTSPO

| Parameter Name | Value |
| --- | --- |
| Amplitude of Fourier (A) | 6.89 |
| Frequency of Fourier (f) | 1.60 |
| Number of Terms of Fourier (n) | 1.00 |
| Phase Shift of Fourier (ϕ) | 0.71 |
| Linear Trend Coefficient (T) | -90.0 |
| Seasonality Amplitude (S) | 45.0 |
| Rates of Change for Frequency (Δf) | -0.27 |
| Rates of Change for Amplitude (ΔA) | 0.57 |
| Rates of Change for Phase Shift (Δϕ) | -0.85 |
| Frequency of Seasonality (SF) | 0.94 |

Secondly, we introduced BOTSPO, which efficiently identifies optimized laser power profiles to enhance heat treatment and precipitation hardening. Given the computational demands of high-fidelity FEM models, BOTSPO is engineered to ascertain optimal profiles with minimal function evaluations. Addressing the potentially infinite dimensionality of time series profiles, we employ a dimension reduction strategy by devising a time series profile generator that leverages a set of 10 parameters, facilitating a comprehensive exploration of the time series space without incurring prohibitive computational costs. This approach is crucial as BO's efficiency inversely correlates with the number of parameters to optimize. We tested this approach with the objective of maximizing the heat treatment time for the thin-wall IN718 build and found that the heat treatment time was maximized to 38.19 seconds. The off-line process optimization result will serve as the reference process trajectory in online model predictive control.

Future work will focus on the integration of the methods presented in this paper with a physical DED system. For physical realization of a DED-DT systems, there are additional tasks to be considered, including online optimization of the laser power profile, continuous model update using both online and off-line data, and drift correction of the surrogate model via model update from a more accurate sub-real-time model.

**Acknowledgements**

The authors would like to acknowledge support from the NSF Engineering Research Center for Hybrid Autonomous Manufacturing Moving from Evolution to Revolution (ERC‐HAMMER) under Award Number EEC-2133630; the National Science Foundation Graduate Research Fellowship under Grant No. DGE-2234667; and fellowship support from The Graduate School at Northwestern University for the Predictive Science & Engineering Design (PS&ED) Cluster project.